\documentclass[runningheads]{llncs}
\usepackage{array,multirow,graphicx}
\usepackage{amsmath}
\usepackage{amssymb}
\usepackage{color, colortbl}
\newcommand{\white}{\textcolor[rgb]{1.0,1.0,1.0}}

\newcommand{\tilof}{\textcolor[rgb]{0.0,0.0,0.0}}

\begin{document}
\title{\tilof{\vspace{-63pt}\\Visual Microfossil Identification\\ via Deep Metric Learning}}
%
%
\author{\tilof{Tayfun Karaderi$^{1,3}$, Tilo Burghardt$^1$, Allison Y. Hsiang$^2$,\\ Jacob Ramaer$^1$, Daniela N. Schmidt$^3$}}
\authorrunning{T Karaderi, T Burghardt, AY Hsiang, J Ramaer, DN Schmidt}
%
\institute{\tilof{$^1$Department of Computer Science, University of Bristol, UK\\
           $^2$Institutionen för geologiska vetenskaper, Stockholm University, Sweden\\
           $^3$School of Earth Sciences, University of Bristol, UK\\}}
\maketitle              
\vspace{-16pt}
\begin{abstract}
\tilof{We apply deep metric learning for the first time to the problem of classifying planktic foraminifer shells on microscopic images. This species recognition task is an important information source and scientific pillar for reconstructing past climates. All foraminifer CNN recognition pipelines in the literature produce black-box classifiers that lack visualisation options for human experts and cannot be applied to open set problems. Here, we benchmark metric learning against these pipelines, produce the first scientific visualisation of the phenotypic planktic foraminifer morphology space, and demonstrate that metric learning can be used to cluster species unseen during training. We show that metric learning outperforms all published CNN-based state-of-the-art benchmarks in this domain. We evaluate our approach on the 34,640 expert-annotated images of the Endless Forams public library of 35 modern planktic foramini-fera species. Our results on this data show leading $92\%$ accuracy (at $0.84$ F1-score) in reproducing expert labels on withheld test data, and $66.5\%$ accuracy (at $0.70$ F1-score) when clustering species never encountered in training. We conclude that metric learning is highly effective for this domain and serves as an important tool towards expert-in-the-loop automation of microfossil identification. Key code, network weights, and data splits are published with this paper for full reproducibility.}

\end{abstract}

\begin{figure}[t]
\begin{center}
\includegraphics[width=1.01\linewidth , height=180pt]{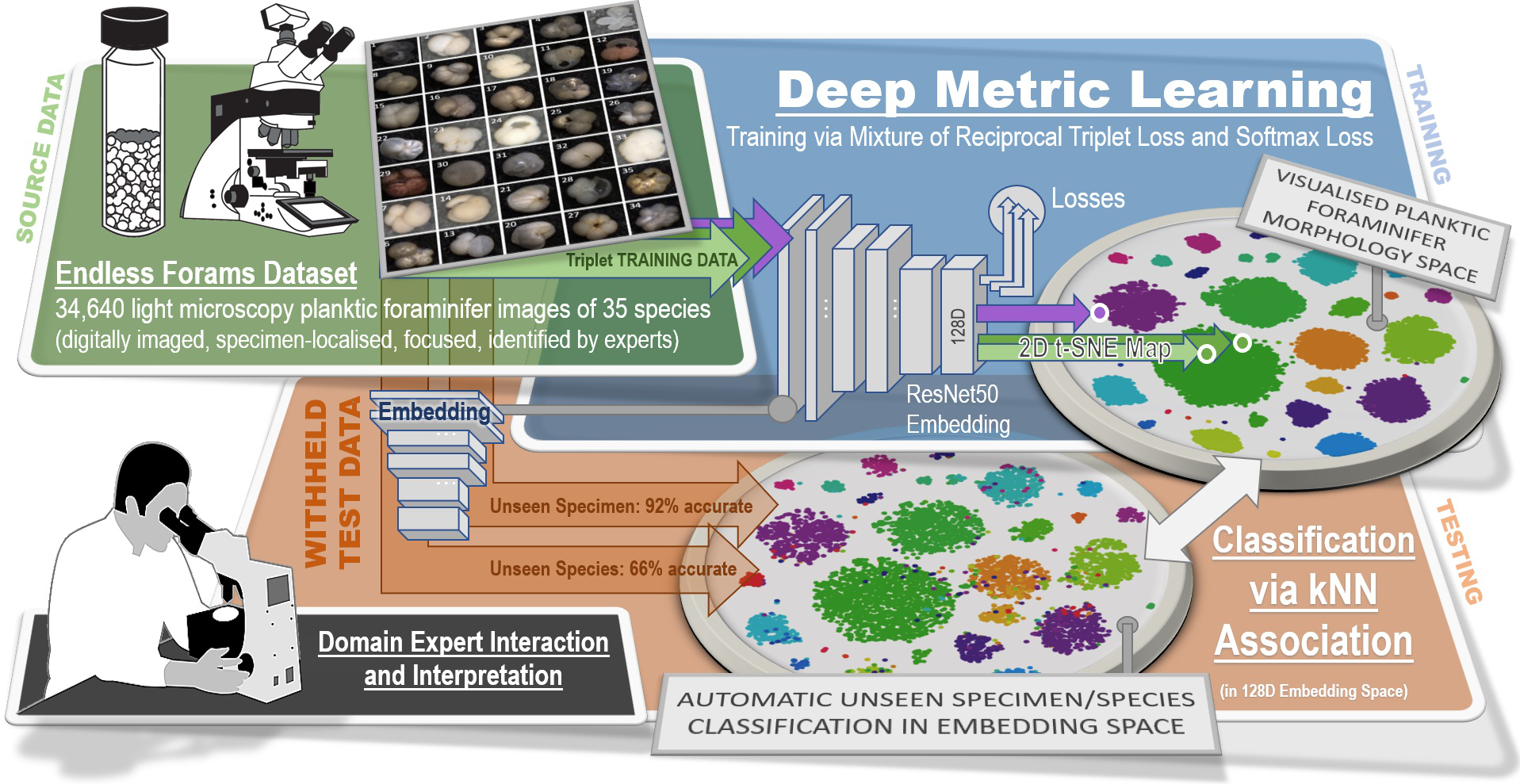}\vspace{-19pt}
\end{center}
   \caption{\tilof{\textbf{Deep Metric Learning of Foraminifer Spaces.} Schematic overview of the approach. Expert-annotated planktic foraminifer imagery (green) is used to metrically learn (blue) embeddings of specimen visuals in a 128D morphological appearance space via a Resnet50-based deep network using hybrid reciprocal triplet and Softmax loss minimisation. The built space naturally allows for a t-SNE visualisation  of specimens and species including their geometric and topological relations. Unseen specimens and even unseen species not used for training can be projected into this space for automatic classification~(red). Domain experts can now for the first time see the location of specimens in the morphological space and interpret results taxonomically. }}\vspace{-11pt}
\label{fig:summary}
\end{figure}
\vspace{-20pt}
\section{Introduction}\vspace{-7pt}
\subsubsection{Motivation.}
\begin{figure}[t]
\begin{center}
\includegraphics[width=1.0\linewidth]{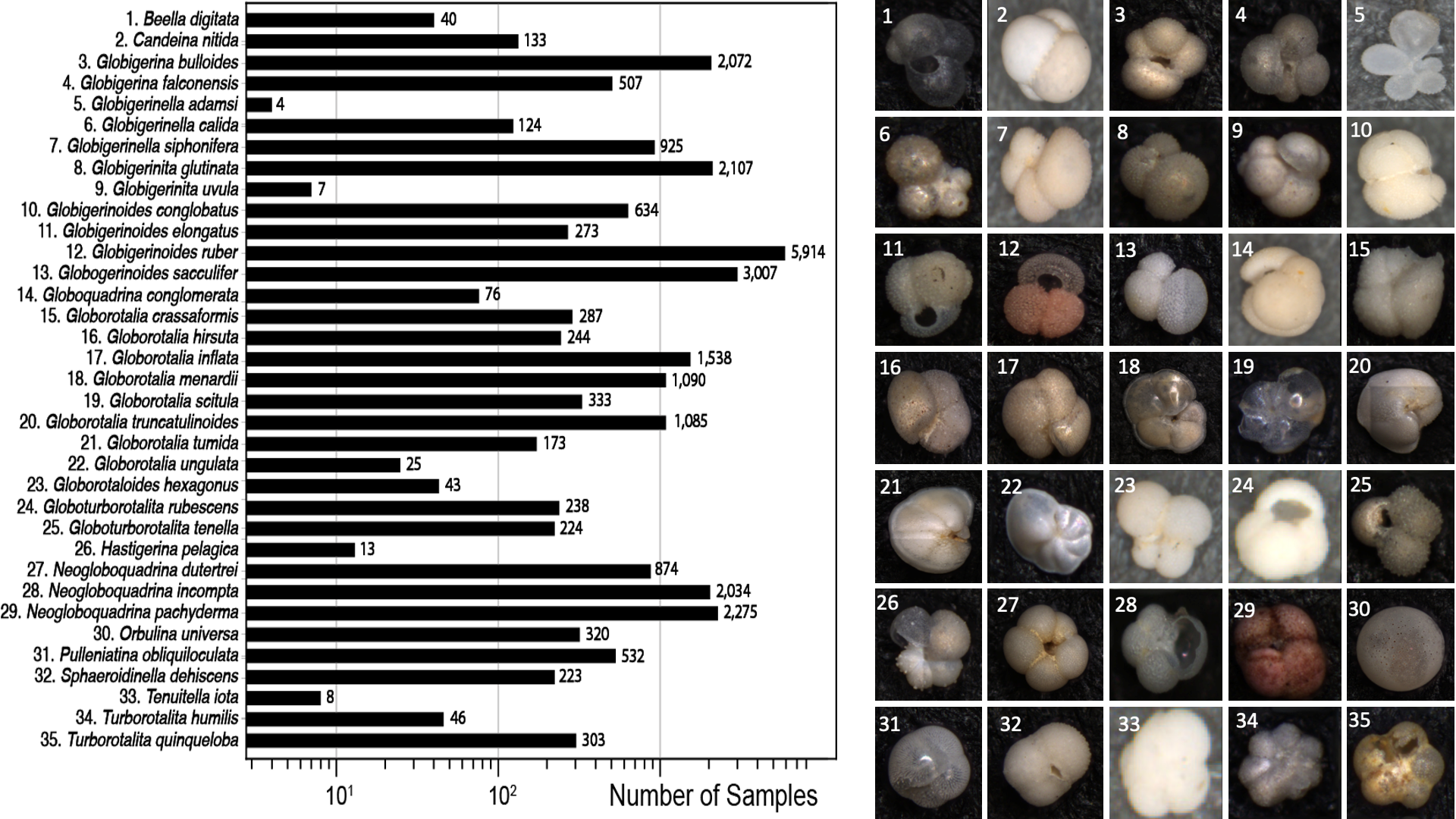}\vspace{-18pt}
\end{center}
   \caption{\tilof{\textbf{Foraminifer Microscope Imagery.} The Endless Forams public image library is used as our data source of modern planktonic foraminifera. \textbf{\textit{(left)}} The distribution of the species classes varies strongly with a mean of 795 and a standard deviation of 1171. \textbf{\textit{(right)}} Visualisation of a sample from each of the 35 species in the dataset.}}\vspace{-8pt}
\label{fig:data}
\end{figure}

\tilof{Planktic foraminifers are an invaluable source of information for reconstructing past climate records~\cite{101}. Estimating ocean temperature, salinity, and pH using foraminifers involves quantifying the species composition of shell assemblages or picking individual specimens, often at low abundance, out of thousands of specimens. Identification of specimens to the species level is necessary as species-specific vital effects can result in different isotopic fractionation values~\cite{RaveloHillaireMarcel}.  Foraminifers grow their calcium carbonate shells by adding chambers in a spiral, where the main gross morphological traits used for taxonomic identification are chamber form and arrangement, the size and position of an aperture, and other features~\cite{102}~(see Figs.~\ref{fig:summary} and~\ref{fig:data}). The differences between species are often plastic along sliding morphological change~\cite{103} and human identifiers manipulate the specimen under the microscope to aid recognition. In contrast, single-view static image recognition confidence can be restricted by acquisition artefacts, imaging quality, and the viewpoint-dependent visibility limitations of traits. These factors pose a significant challenge with regard to the use of computer vision systems for automating single image identification.} \vspace{-12pt}
\subsubsection{Paper Contributions.}
\tilof{Inspired by classic machine learning~(ML) applications for microfossils~\cite{Balfoort,Beaufort2004,Weller}, recent pioneering works~\cite{2,marchant20} started to evaluate convolutional neural networks (CNNs) for visual foraminifer recognition. However, none of the published pipelines so far allow for meaningful visualisations and expert interactions with the learned space of specimens, application to unseen species, or utilisation of contrastive gradients. In this paper, we address the above limitations for the first time via deep metric learning~\cite{andrew21,schneider19} in the domain. Fig.~\ref{fig:summary} illustrates our approach visually. After a review of background and methodology, we will experimentally demonstrate that metric learning can outperform the current state-of-the-art~(SOTA) in foraminifer classification benchmarks. In addition, we will show that our approach can generate grey-box foraminifer classification models with which domain experts can interact. Finally, we will explore how far the projection of unseen foraminifer species into this space (zero-shot scenario) can generalise learning beyond trained foraminifer classes.}\vspace{-10pt}
  
\section{Background}\vspace{-7pt}
\subsubsection{Manual Microfossil Identification.}
\tilof{Despite the importance of foraminifers in paleoclimatology, few aggregated public resources exist to train people in the task of distinguishing {morphologically similar} taxa. Further, diverging views on species concepts and boundaries (e.g., `clumpers' vs. `splitters') result in conflicting taxonomies in the literature. As a result, the taxonomic agreement is sometimes only $\sim$ 75\%~\cite{13} for planktic foraminifera since they exhibit near‐continuous morphological gradations between closely related taxa~\cite{14}~(see Fig.~\ref{fig:data}). In some cases, morphological variation is unrelated to genetic differentiation; however in others genetic analysis has revealed the existence of some pseudo‐cryptic species between morphological endmembers~\cite{16}. These difficulties resulted in different species concepts over time, which are particularly prominent in self‐trained taxonomists~\cite{13}. Some of these challenges might be removed by growing databases of expert-classified images~\cite{301}, opening opportunities for machine-driven classification that limit subjective biases of human classifiers.}\vspace{-15pt}
\subsubsection{Current Machine Vision for Microfossil Classification.} \tilof{Modern semantic image classification frameworks are almost without exception grounded in feed-forward CNNs, introduced in their earliest form by the ground-breaking work of Krizhevsky et al.~\cite{cnn} and leading to further milestone architectures including VGG16~\cite{vgg}, Inception~\cite{inception}, ResNet~\cite{resnet}, ResNeXt~\cite{resnext}, and deep transformer networks~\cite{deeptransformer}. Taxonomic computer vision (CV) applications of these techniques for microfossils are still very rare in the literature, despite steep advances in general animal biometrics~\cite{kuhl,tuia}. Nevertheless, ML-based species identification~\cite{mlp1,mlp2} has been applied to several microscopic taxa, including coccoliths~\cite{5}, pollen~\cite{6}, and phytoplankton~\cite{7}. While there have been early attempts for automatic classification on marine microfossils~\cite{Balfoort,Weller} the most successful focused on coccoliths which are predominantly flat ~\cite{Beaufort2004,Beaufort2011}. There exist only very few papers which investigate the use of modern deep learning techniques on planktic foraminifers~\cite{2,marchant20,mitra}, all of which put forward traditional non-contrastive CNN architectures optimising for prediction correctness via SoftMax cross-entropy.}\vspace{-6pt}
\subsubsection{Metric Latent Spaces.}\tilof{Metric learning~\cite{andrew21,schneider19} moves away from focusing learning on optimising prediction correctness only; instead, a mapping into a class-distinctive latent space is constructed where maps to the same class naturally cluster together and distances directly relate to input similarity under the training task. A simple way of building a latent space of this form is to pass two inputs through an embedding function and then use a contrastive loss~$L_{C}$~\cite{20}:}
\begin{equation}
L_{C} = (1-Y)0.5d(x_{1}, x_{2})+0.5Ymax(0, \alpha - d(x_{1}, x_{2})),
\end{equation}

\noindent
\tilof{where $x_{1}$ and $x_{2}$ are the embedded input vectors, $Y$ is a binary label denoting class equivalence/difference for the two inputs, and $d(\cdot,\cdot)$ is the Euclidean distance between two embeddings. However, this formulation cannot put similarities and dissimilarities between different pairs of embeddings in relation. A triplet loss formulation~\cite{21} instead utilises embeddings $x_a$, $x_p$ and $x_n$ denoting an anchor, a positive example of the same class, and a negative example of a different class, respectively. Minimising the distance between the same-class pair and maximising the distance between the different-class pair can be achieved by:}
\begin{equation}
L_{TL} = max(0; d(x_a,x_p) - d(x_a,x_n) + \alpha), 
\end{equation}
\tilof{where $\alpha$ is the margin hyper-parameter. Reciprocal triplet loss removes the need for this parameter~\cite{22} and accounts for large margins far away from the anchor:}
\begin{equation}
L_{RTL} = d(x_a,x_p) +1/d(x_a,x_n).
\end{equation}
\tilof{Including a SoftMax term in this loss can improve performance, as shown by recent work~\cite{23,24}. Thus, the SoftMax and reciprocal triplet losses can be combined into the standard formulation first published in~\cite{andrew21} used in this paper:}
\begin{equation}
L = \frac{-log(e^{x_{class}})} {\sum_i e^{x_{i}}} + \lambda L_{RTL},
\end{equation}
\tilof{where $\lambda$ is a mixing hyper-parameter. For the foraminifer classification problem in particular this allows both relative inter-species difference information captured by the reciprocal triplet loss component as well as overall species information captured by the SoftMax term to be used as backpropagation gradients.}\vspace{-8pt} 
\subsubsection{Latent Space Partitioning for Classification.}
\begin{figure}[t]
\begin{center}
   \includegraphics[width=1\linewidth, height=112pt]{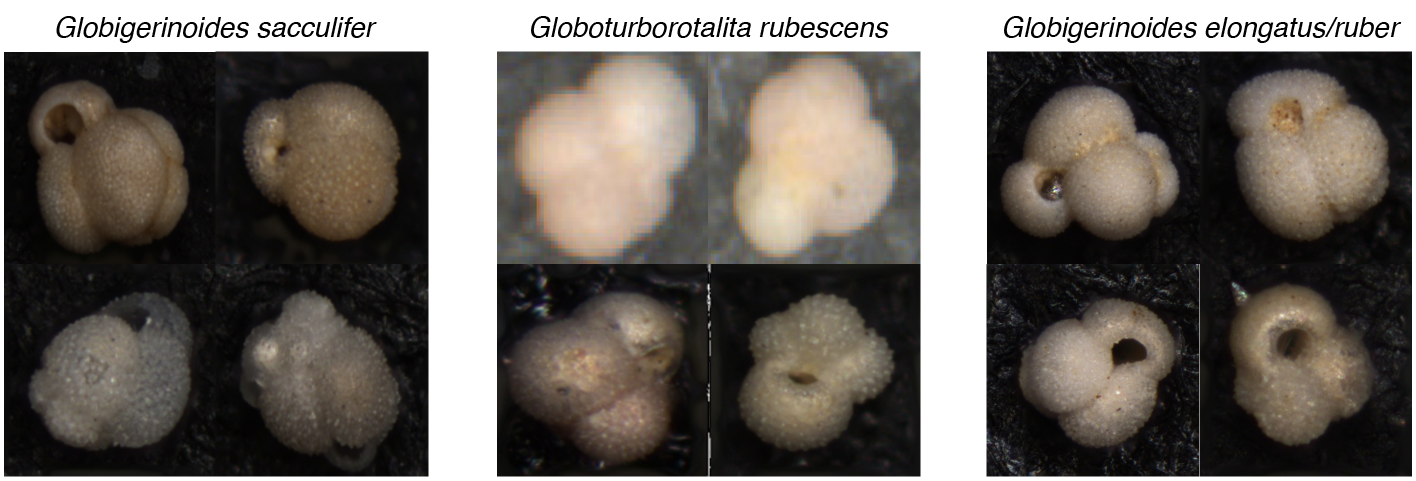}\vspace{-17pt}
\end{center}
   \caption{\tilof{\textbf{Imaging Details and Classification Challenges.} Microscopic imaging of 3D microfossils often obfuscates key taxonomic features due to variable viewpoint and acquisition conditions. \textbf{\textit{(left)}}~Within-class 3D viewpoint variability for \textit{Globigerinoides sacculifer} yields 2D images with different visible chamber numbers rendering this key feature less informative. \textbf{\textit{(middle)}}~Out-of-focus acquisition can remove species-characteristic texture information, here of \textit{Globoturborotalita rubescens}, blurred in the top images. Light microscopy images are also unable to capture fine grain wall texture details that would be useful for taxonomic identification. \textbf{\textit{(right)}}~{Morphologically similar species such as} \textit{Globigerinoides elongatus} (top) and \textit{G. ruber} (bottom) are {difficult to distinguish} from {single-view static} visuals alone. Ground truth labeling often requires additional information to establish secure taxonomic classification.}}\vspace{-5pt}
\label{fig:datadetail}
\end{figure}
\tilof{Once an embedding function has been learned for a metric space, any new sample -- even one of the unseen classes -- can be projected into this space. However, assigning a class label to this sample based on its position in the space eventually requires a partitioning of the latent space. Direct maps, as well as hierarchical~\cite{hcs} and partitional~\cite{partition_review} clustering algorithms, have been used for this. We will experiment with a wide variety of partitioning options in Tab.~\ref{table:resolution}\textit{(top right)} including Gaussian Mixture Models (GMMs)~\cite{gmms}, Logistic Regression, Support Vector Machines (SVMs)~\cite{svm}, Multi-layer Perceptrons (MLPs)~\cite{mlp1}, and k-Nearest Neighbours~\cite{knn}. Once partitioned, metric spaces expose model structure plus outlier information, can be visualised by mapping into lower-dimensional spaces, and often capture properties of the target domain beyond the specific samples used for training. Before using these techniques on the problem of foraminifer classification, we will first introduce the dataset and experimental settings.}\vspace{-5pt}

\section{Dataset}\vspace{-7pt}
\subsubsection{Endless Forams.}\tilof{We use the Endless Forams image library~\cite{1} for all experiments. It is one of the largest datasets of its kind and publically available (at \url{endlessforams.org}). It contains 34,640 labelled images of 35 different foraminiferal species as detailed and exemplified in Fig.~\ref{fig:data}. This dataset was built based on a subset of foraminifer samples from the Yale Peabody Museum (YPM) Coretop Collection~\cite{elder} and the Natural History Museum, London (NHM) Henry A. Buckley Collection~\cite{rillo}. The dataset is also associated with a taxonomic training portal hosted on the citizen science platform Zooniverse~(\url{zooniverse.org/projects/ahsiang/endless‐forams}). Species classification in this dataset is truly challenging compared to many other computer vision tasks since: 1) planktonic foraminifers exhibit significant intra-class variability (see Fig.~\ref{fig:datadetail}\textit{(left)}); 2) critical morphological properties are not consistent across 3D viewpoint and acquisition conditions (see Fig.~\ref{fig:datadetail}\textit{(left)}, \textit{(middle)}); and 3) visual intra-species differences are barely apparent between some taxa (see Fig.~\ref{fig:datadetail}\textit{(right)}).}\vspace{-7pt}

\begin{figure}[t]
\begin{center}
\includegraphics[height=130pt]{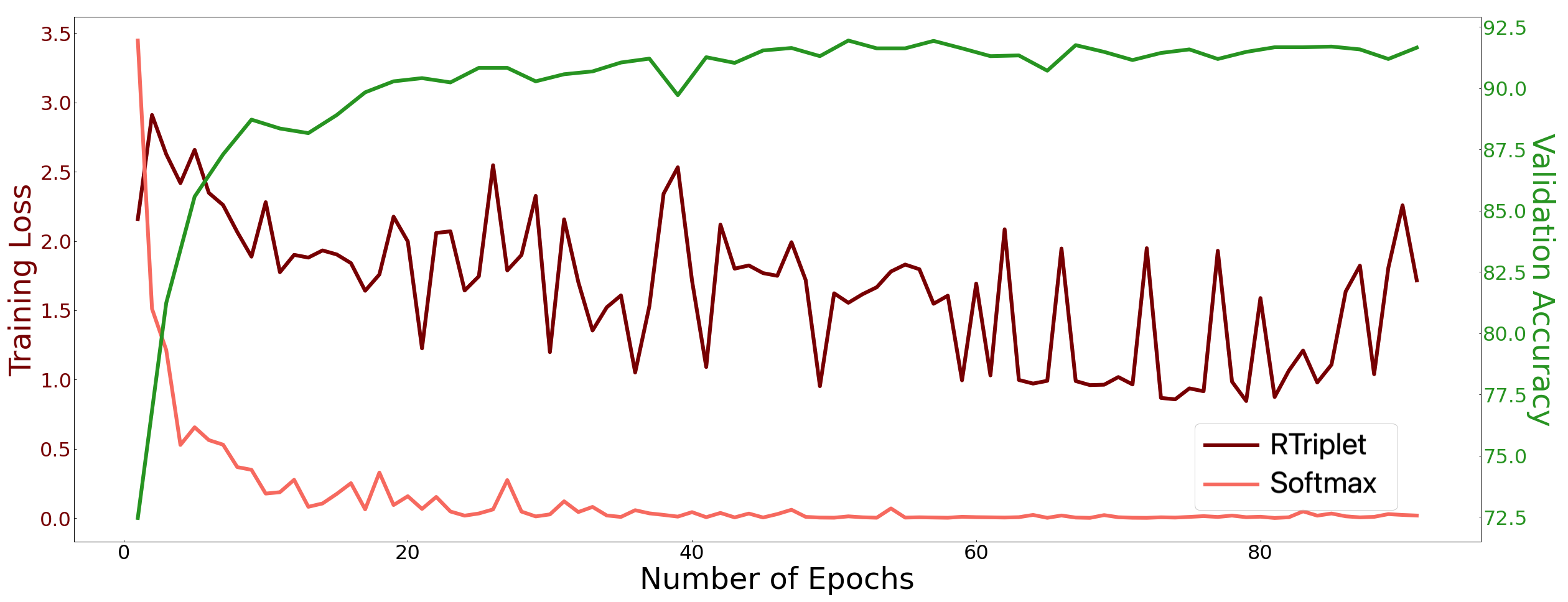}\vspace{-15pt}
\end{center}
   \caption{\tilof{\textbf{Metric Learning Training Details.} Curves quantifying the evolution of training losses and validation accuracy across all 100~epochs of training. Note that the two loss components are plotted separately showing the early conversion of the Softmax dimension, whilst RTriplet loss continues to alter the metric embedding space at higher magnitudes albeit at down-scaled weighting given $\lambda=0.01$ in Eq.~4. }}\vspace{-5pt}
\label{fig:train}
\end{figure}
\section{Experimental Setup}\vspace{-7pt}
\label{sec-imp}
\subsubsection{Implementation Details.} \tilof{For all experiments, our PyTorch-implemented metric learning architecture extends a ResNet50 backbone pre-trained with ImageNet. The network is set to optimise the loss specified in Eq.~4 combining SoftMax and reciprocal triplet loss components with the mixing parameter $\lambda=0.01$ as suggested in~\cite{24}.  Training progresses via the SGD~\cite{sgd} optimiser for 100~epochs as quantitatively illustrated in~Fig.~\ref{fig:train}. For full testing comparability with~\cite{1} we utilised their withheld test set for performance stipulation, whilst using the remaining 27,731 images augmented via rotations, scale, and Gaussian noise transforms for training~(reported as \textbf{Ours}). In a second workflow and for full compatibility with~\cite{marchant20}, we also produced results via 5-fold cross-validation on random train-test data splits~(reported as \textbf{Ours*}). For sample selection during training of all workflows, we follow the ‘batch hard’ mining approach~\cite{batch_hard} where triplets are mined within mini-batches. This yields overall moderate triplets, i.e. training with the hardest examples within each mini-batch. The published source code~\cite{repo} provides full details regarding all of the above for full result reproducibility. Training takes approx. 48 hours on a P100 GPU system with 12GB RAM. We obtain classifications for each test sample projected into the metric space via running kNN~\cite{knn} with n=5 over the projected training samples.}\vspace{-7pt} 

\begin{table}[t]
\scriptsize{
\begin{center}
\begin{tabular}{|c|c|l|c|c|c|c|c|}

\hline
\white{.........}& \white{...}\textbf{Row}\white{...} & \textbf{Method}  & \textbf{Resolution} & \textbf{Acc}  & \textbf{Prec}   & \textbf{Rec}  & \textbf{F1S} \\ 
 & &  & (pixels) & (weighted) & (macro) & (macro) & (macro) \\ \hline\hline
\parbox[t]{2mm}{\multirow{6}{*}{\rotatebox[origin=c]{0}{A}}} 
& 01 & VGG19*~\cite{marchant20} & $224\times 224$ & 77.1 &70.5 & 64.9 &66.9 \\ \cline{2-8}
& 02 & MobileNetV2*~\cite{marchant20}  & $224\times 224$ & 77.7 & 70.0 & 65.2  & 66.8\\ \cline{2-8}
& 03 & InceptionV2*~\cite{marchant20}  & $224\times 224$ & 77.7 & 69.5 & 64.8  & 66.4\\ \cline{2-8}
& 04 & DenseNet121*~\cite{marchant20}  & $224\times 224$ & 80.2 & 75.2 & 69.2  & 71.3 \\ \cline{2-8}
& 05 & ResNet50*~\cite{marchant20}  & $224\times 224$ & 81.8 & 76.7 & 71.4 & 73.4\\ \cline{2-8}
& 06 & \textbf{Ours*}& $224\times 224$ & \textbf{91.6} & \textbf{88.2} & \textbf{78.3} & \textbf{81.3} \\ \hline\hline

\parbox[t]{2mm}{\multirow{2}{*}{\rotatebox[origin=c]{0}{B}}} 
& 07 & VGG16~\cite{1} & $160\times 160$ & 87.4 & 72.4 & 69.8 & 70.0 \\ \cline{2-8}
& 08 & \textbf{Ours} & $160\times 160$ & \textbf{91.9} & \textbf{91.3} & \textbf{81.9}  & \textbf{84.6} \\ \hline\hline

\parbox[t]{2mm}{\multirow{4}{*}{\rotatebox[origin=c]{0}{C}}} 
& 09 & ResNet18Full*~\cite{marchant20} & $128\times 128$ & 88.5 &84.1 & 77.8 & 79.9 \\ \cline{2-8}
& 10 & ResNet50CycleFull*~\cite{marchant20}\white{.........} & $128\times 128$ & 90.1 &85.1 &78.7  & 80.8 \\ \cline{2-8}
& 11 & BaseCycleFull*~\cite{marchant20} & $128\times 128$ & 90.3 & 84.9& 78.4 &80.5 \\ \cline{2-8}
& 12 & \textbf{Ours*} & $128\times 128$ & \textbf{91.0} & \textbf{87.2} & \textbf{78.9}  & \textbf{81.1} \\ \hline\hline

\parbox[t]{2mm}{\multirow{3}{*}{\rotatebox[origin=c]{0}{D}}} 
& 13 & BaseCycleFull~\cite{marchant20} & $224\times 224$ & 90.5 & 84.5 & 79.6  & 81.5 \\ \cline{2-8}
& 14 & \textbf{Ours} & $224\times 224$ & \textbf{91.7} & \textbf{88.1} & \textbf{77.0} & \textbf{80.7} \\ \cline{4-8}
&  &  & $416\times 416$ & \textbf{92.0} & \textbf{89.0} & \textbf{81.5}  & \textbf{84.2}\\ \hline

\end{tabular}
\end{center}}~\vspace{-10pt}
\caption{\textbf{Species Classification Results.} \tilof{Performance of our metric learning approach against SOTA techniques~\cite{marchant20,1} quantified via accuracy, precison, recall and F1 score. Testing regime and input resolution were made to match the settings of  previous works exactly where \textbf{*} indicates 5-fold cross-validation~(see Sec.~\ref{sec-imp}). \textbf{\textit{(A)}}~Metric learning supremacy against various off-the-shelf CNNs at $224\times 224$ initialised via ImageNet as described in~\cite{marchant20}. \textbf{\textit{(B)}}~Metric learning increases accuracy of~\cite{1} at $160\times 160$ by $4.5\%$. \textbf{\textit{(C)}}~Critically, our approach outperforms SOTA CycleNet~\cite{marchant20} at their published $128\times 128$ resolution. \textbf{\textit{(D)}}~Finally, re-running CycleNet~\cite{marchant20} under the regime of~\cite{1} and at $224\times 224$ shows metric learning also dominating peaking at $92\%$ at $416\times 416$.}}~\vspace{-25pt}
\label{table:results}
\end{table}

\label{sec_results}

\section{Results.}~\vspace{-38pt}
\subsubsection{Baseline Comparisons.} \tilof{The simplest baselines for the foraminifer species classification problem are given by using transfer learning via ImageNet-initialised off-the-shelf CNN architectures~\cite{marchant20}. Tab.~\ref{table:results}\textit{(rows 01-06)} compares key architectures and their performance against our setup using 5-fold cross-validation at standard resolution of $224\times 224$. These results show that metric learning dominates such baselines without exception. Operating at $160\times 160$ pixels, the original 2019 VGG-based benchmark by Hsiang et al.~\cite{1} achieved~$87.4\%$ accuracy on a withheld test set. Using the same test set, our metric learning approach improves on this performance significantly by $4.5\%$ as shown in Tab.~\ref{table:results}\textit{(rows 07-08)}.}~\vspace{-16pt}
\subsubsection{Improving the State-of-the-Art.} \tilof{CycleNet used by Marchant et al.~\cite{marchant20} claims current state-of-the-art performance on the data at an accuracy of~$90.3\%$  when using  5-fold cross-validation at $128\times 128$. Tab.~\ref{table:results}\textit{(rows 09-12)} compare their key results against our metric learning approach under this regime, showing improved metric learning accuracy of $91\%$. When using their best-performing CycleNet~\cite{marchant20}, scaling it up to a standard $224\times 224$ resolution and using the testing regime of~\cite{1} their results improve slightly. However, our approach outperforms this new benchmark by another $1.2\%$ as shown in Tab.~\ref{table:results}\textit{(rows 13-14)}. Metric learning reaches top accuracy of~$92\%$ at resolution $416\times 416$ as shown in Tab.~\ref{table:results}\textit{(row~14)}. Fig.~\ref{fig:latent}\textit{(right)} depicts a detailed confusion matrix for this setting.}~\vspace{-18pt}
\subsubsection{Visualisation of Foraminifer Space.} 
\begin{figure}[t]
\begin{center}
   \includegraphics[width=207pt]{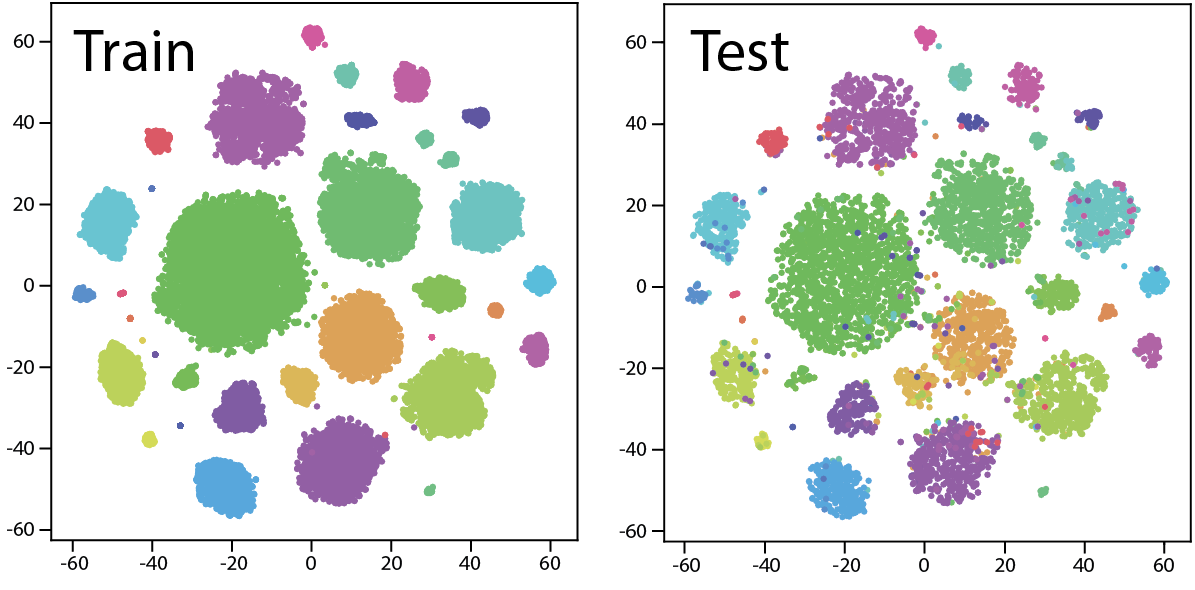}\white{..}\includegraphics[width=139pt]{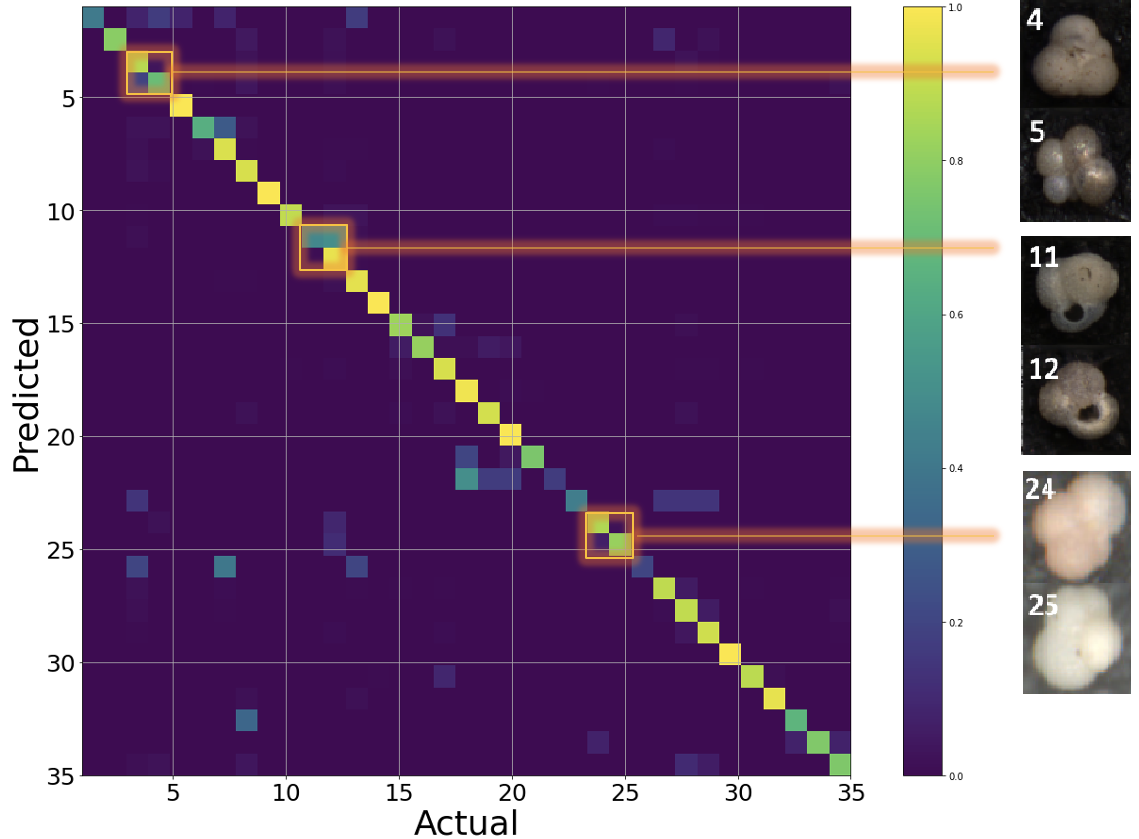}\vspace{-19pt}
\end{center}
   \caption{\tilof{\textbf{Learned Metric Space Visualisation and Confusion Matrix.} \textbf{\textit{(left)}} 2D t-SNE visualisation of the training (RandIndex=$99.9$) and testing (RandIndex=$87.0$) sets projected into the metric space; and \textbf{\textit{(right)}} Confusion matrix detailing test performance where taxonomic similarities of prominent mix-ups are visually highlighted.}}\vspace{-5pt}
\label{fig:latent}
\end{figure}

\tilof{In contrast to basic CNN approaches, we can now visualise the learned 128D metric space, revealing an appearance-based distribution of planktic foraminifers. Fig.~\ref{fig:latent}\textit{(left)} shows this first scientific visualisation of phenotypic appearance space by projecting training and testing sets into the metric space before using t-SNE~\cite{t_sne} to reduce dimensionality to~2.}~\vspace{-18pt}
\subsubsection{Open Set Performance.} \tilof{By withholding some species\footnote{Tail classes 1, 5, 9, 14, 22, 23, 26, 29, 33, and 34 were chosen as our open set to have maximum specimen counts available during training.} from training altogether, we can start to evaluate metric learning potential beyond seen training species, i.e. learning about the planktic foraminifer domain more generally. To do this, we project the unused training data of withheld classes during testing into the metric space together with the training data of used classes. We then utilise kNN with all these data points to measure how far test sets of seen or unseen classes are classified correctly. For the unseen species classes at $224\times 224$, accuracy drops to $66.5\%$ (at $0.70$ F1-score) in our experiment. This result is remarkable since about $2/3$ of never-seen species specimens can still be associated correctly amongst all 35 species. We thus conclude that metric learning does indeed capture important general features relevant for planktic foraminifer classification beyond any particular species appearance.}~\vspace{-9pt}

\subsubsection{Ablations, Resolution, and Augmentation.} \tilof{Tab.~\ref{table:resolution}\textit{(top)} details ablation experiments that demonstrate that the chosen metric learning loss functions and clustering algorithms are effective, contribute to performance, and indeed outperform other tested setups. An analysis of the dependency of our approach on resolution is presented in Tab.~\ref{table:resolution}\textit{(bottom left)}. It outlines that above $160\times 160$ pixels performance gains are widely diminished and flat-line, supporting the choices taken in~\cite{1}. Whilst SOTA competitor networks~\cite{marchant20} also augment their data, augmentation can be essential for metric learning techniques generally~\cite{schneider19}. Tab.~\ref{table:resolution}\textit{(bottom right)} quantifies this fact and shows that performance in the foraminifer classification domain is indeed intricately linked to augmentation, with rotation variations producing the strongest component impact.}~\vspace{-13pt}

\section{Taxonomic Reflection.}~\vspace{-39pt}
\subsubsection{Taxonomic Interpretation of Performance.} \tilof{We observe that species that are taxonomically related to each other are indeed the ones that are confused most often by the machine. For example, species from the same genus (e.g. 4 vs 5; 11 vs 12;    24 vs 25) are often misclassified symmetrically as highlighted in Fig.~\ref{fig:latent}\textit{(right)}. This alignment is consistent with human classification difficulties and suggests that metric space distances are indeed related to visual properties humans use to differentiate species. Generally, we suggest three main reasons for lower classification performances, all of which would also present challenges for human classifiers. First, phylogenetic closeness, i.e. closely related sister taxa, can result in many shared taxonomic features that make differentiation difficult. Classes 4 and 5, for example, share many large-scale morphological features and are only distinguished by a lip at the aperture~\cite{201}. Secondly, some classes (e.g. 11 and 12) are not even typically distinguished in palaeontological studies, but are genetically distinct~\cite{203} and hence separated in the database. The distinguishing visual feature is a sliding scale of radial chamber compression which makes systematic labelling challenging. Thirdly, not all planktic foraminiferal species reach the same maximum size in the modern ocean~\cite{204}, and size is an important feature used for taxonomic classification by human experts. As all the images have the same number of pixels, size information is lost in the database used and the amount of information per pixel will vary strongly between a species which is up to 5 times larger than another. For the smallest specimens, imaging often reaches the limits of typical optics, resulting in blur and other effects, as shown in Fig.~\ref{fig:datadetail}\textit{(middle)} for class 24.}~\vspace{-15pt}

\section{Conclusion.}~\vspace{-43pt}
\subsubsection{}\tilof{The accurate and efficient visual recognition of planktic foraminifers within shell assemblages in light microscopy imagery is an important pillar required for unlocking climatic archives. We have shown here for the first time that deep metric learning can be effectively applied to this task. We documented how deep metric learning outperforms all published state-of-the-art benchmarks in this domain when tested on one of the largest public, expert-annotated image libraries covering modern planktic foraminifera species. We further provided detailed result evaluation and ablation studies. Based on the metrically-learned feature spaces, we also produced the first scientific visualisation of a phenotypic planktic foraminifer appearance domain. Finally, we demonstrated that metric learning can be used to recognise specimens of species unseen during training by utilising the metric space embedding capabilities.  We conclude that metric learning is highly effective for this domain and can form an important tool towards data-driven, expert-in-the-loop automation of microfossil identification.}~\vspace{-15pt}

\section*{Acknowledgements.}~\vspace{-43pt}
\subsubsection{}\tilof{TK was supported by the UKRI CDT in Interactive Artificial Intelligence under the grant EP/S022937/1. AYH was supported by VR grant 2020-03515. DNS was supported by NERC grant NE/P019439/1. We thank R Marchant and his team for making available source code and testing regime details to compare to \cite{marchant20}. Thanks to M Lagunes-Fortiz and W Andrew for permitting use and adaptation of source code related to metric learning.}

\begin{table}[t]
\scriptsize{
\begin{center}
\begin{tabular}{|c|c|c|c|c|c|c|c|c|c|c|c|c|}

\cline{1-6} \cline{8-13}
\textbf{Loss} & \textbf{Acc}  & \textbf{Prec}   & \textbf{Rec}  & \textbf{F1S} & \textbf{RI}& \white{....} & \textbf{Alg} & \textbf{Acc}  & \textbf{Prec}   & \textbf{Rec}  & \textbf{F1S} & \textbf{RI}\\ \cline{1-6} \cline{8-13}

\textbf{Ours} ($L_{RTL} + SM$) & \textbf{91.9} & \textbf{91.3} & \textbf{81.9}& \textbf{84.6}& \textbf{87.1} &  & \white{..}\textbf{Ours} (kNN)\white{.} & \textbf{91.9} & \textbf{91.3} & \textbf{81.9} &\textbf{84.6}  & \textbf{87.1}\\ \cline{1-6} \cline{8-13}
SoftMax+$L_{TL}$ & 91.4 & 89.6 & 81.3 & 84.3 & 86.2 & & SVM & { 91.8 } & 88.3 & 78.8 & 81.6 & 86.6\\ \cline{1-6} \cline{8-13}
\white{..}RTriplet ($L_{RTL}$)\white{..} & 89.3 & 79.2 & 71.0 & 73.9 & 82.9 & & MLP & { 90.9 } & 80.9 & 80.8 & 78.9 & 86.2\\ \cline{1-6} \cline{8-13}
Triplet ($L_{TL}$) & 22.2 & 12 & 9.3 & 9.5 & 4.7 & & LR & { 91.5 } & 84.6  & 74.7& 78.0 & 86.4  \\ \cline{1-6} \cline{8-13}
SoftMax & 91.2 & 89.9 & 81.9& 84.2& 86.5 & & GMM & { - } & - & - & - & 63.8 \\ \cline{1-6} \cline{8-13}

\end{tabular}\vspace{7pt}
\begin{tabular}{|c|c|c|c|c|c|c|c|c|c|c|c|c|}

\cline{1-6} \cline{8-13}
\white{..}\textbf{Resolution} \white{..}& \textbf{Acc}  & \textbf{Prec}   & \textbf{Rec}  & \textbf{F1S} & \textbf{RI}& \white{....} & \textbf{Transform} & \textbf{Acc}  & \textbf{Prec}   & \textbf{Rec}  & \textbf{F1S} & \textbf{RI}\\ \cline{1-6} \cline{8-13}

$416\times 416$ & \textbf{92.0} & 89.0 & 81.5 & 84.2 & \textbf{87.3} & &

\white{...}\textbf{Ours} (R+S+G)\white{....} & \textbf{91.9} & \textbf{91.3} & \textbf{81.9}& \textbf{84.6}& \textbf{87.1}\\ \cline{1-6} \cline{8-13}

$224\times 224$ & 91.7 & 88.1 & 77.0 & 80.7 & 86.8 & & R & {91.2} &86.8 &79.3 & 81.5& 85.8\\ \cline{1-6} \cline{8-13}

$160\times 160$ & 91.9 & \textbf{91.3} & \textbf{81.9} & \textbf{84.6} & 87.1 & & S & {88.4} & 87.6 & 76.3 & 80.0 & 81.1  \\ \cline{1-6} \cline{8-13}

$120\times 120$ & 91.0 & 88.4 & 79.2 & 82.1&86.3 & & G & {85.0} & 75.0 &66.2 &69.1 &76.3  \\ \cline{1-6} \cline{8-13}

$80\times 80$ & 90.0 & 84.8 & 78.1 & 80.2 & 84.1 & & -- & {82.7} &75.2 &66.2 &69.0 &72.4  \\ \cline{1-6} \cline{8-13}

\end{tabular}
\end{center}}~\vspace{-10pt}
\caption{\textbf{Design Ablations, Resolution and Augmentation.} \tilof{Benchmarks use the testing regime of~\cite{1} and resolution $160\times 160$ as standard; RI denotes the Rand Index. \textbf{\textit{(top left)}}~Performance of different loss functions confirm the superiority of $L$. \textbf{\textit{(top right)}}~Benchmark of various clustering approaches. \textbf{\textit{(bottom left)}}~Performance increases saturate around $160\times 160$ providing only marginal gains if input resolution is increased beyond this point. \textbf{\textit{(bottom right)}}~Metric learning performance in this domain is intricately dependent on sufficient data augmentation. Rotation augmentation~(R) in particular benefits accuracy, whilst scale augmentation~(S), and Gaussian noise addition~(G) have smaller effects. Applying all three (R+S+G) is most beneficial.}}~\vspace{-25pt}
\label{table:resolution}
\end{table}

\bibliographystyle{splncs04} 
\tiny
\bibliography{samplepaper.bib}

\end{document}